%% file: main.tex
\documentclass[10pt,twocolumn,letterpaper]{article}

\usepackage{iccv}              
\usepackage{xcolor}
\usepackage{multirow}
\usepackage{pifont}
\usepackage{colortbl, booktabs}
\usepackage{array}

\definecolor{iccvblue}{rgb}{0.21,0.49,0.74}
\usepackage[pagebackref,breaklinks,colorlinks,allcolors=iccvblue]{hyperref}
\def\paperID{9524} 
\def\confName{ICCV}
\def\confYear{2025}

\title{Player-Centric Multimodal Prompt Generation for Large Language Model Based Identity-Aware Basketball Video Captioning}

\author{Zeyu Xi{$^{1}$},~ Haoying Sun{$^{1}$},~ Yaofei Wu{$^{1}$},~ Junchi Yan{$^{2}$},~ Haoran Zhang{$^{1}$},~ Lifang Wu{$^{1*}$},\\~ Liang Wang{$^{3}$},~ Changwen Chen{$^{4}$}\\\vspace{-8pt}{\small~}\\
{$^{1}$}Beijing University of Technology;~~~ {$^{2}$}Shanghai Jiao Tong University;~~~ {$^{3}$}Chinese Academy of Sciences \\ {$^{4}$}The Hong Kong Polytechnic University\\
{$^{*}$}Corresponding author: lfwu@bjut.edu.cn}

\begin{document}
\maketitle

\input{all_section}

{
    \small
    \bibliographystyle{ieeenat_fullname}
    \bibliography{main}
}

\input{appendix}

\end{document}

%% file: all_section.tex
\begin{abstract}
Existing sports video captioning methods often focus on the action yet overlook player identities, limiting their applicability. Although some methods integrate extra information to generate identity-aware descriptions, the player identities are sometimes incorrect because the extra information is independent of the video content. This paper proposes a player-centric multimodal prompt generation network for identity-aware sports video captioning (LLM-IAVC), which focuses on recognizing player identities from a visual perspective. Specifically, an identity-related information extraction module (IRIEM) is designed to extract player-related multimodal embeddings. IRIEM includes a player identification network (PIN) for extracting visual features and player names, and a bidirectional semantic interaction module (BSIM) to link player features with video content for mutual enhancement. Additionally, a visual context learning module (VCLM) is designed to capture the key video context information. Finally, by integrating the outputs of the above modules as the multimodal prompt for the large language model (LLM), it facilitates the generation of descriptions with player identities. To support this work, we construct a new benchmark called NBA-Identity, a large identity-aware basketball video captioning dataset with 9,726 videos covering 9 major event types. The experimental results on NBA-Identity and VC-NBA-2022 demonstrate that our proposed model achieves advanced performance. Code and dataset are publicly available at \url{https://github.com/Zeyu1226-mt/LLM-IAVC}.
\end{abstract}

\section{Introduction}
\label{sec:intro}

Video captioning is a critical research topic in video understanding with the aims of generating precise descriptions to comprehend the visual contents of the given videos. It has attracted increasing attention due to a variety of its applications including sports video understanding \cite{xi2024knowledge, qi2023goal, rao2024matchtime, mkhallati2023soccernet, wu2022sports}, video storytelling \cite{Storytelling} and online video search \cite{DBLP:conf/mm/NieQM00B22, DBLP:journals/corr/abs-2309-11091}.

\begin{figure}[t!]
\centering
\includegraphics[width=8.4cm,height=4.0cm]{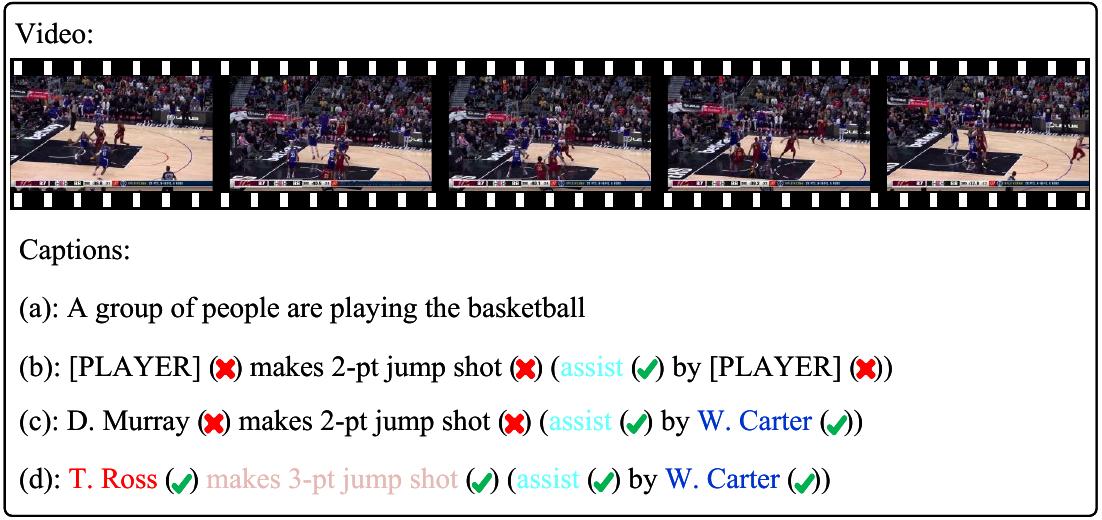}
	\caption{Caption comparison of current sports video captioning methods and our method. The different player names are marked {\color{red}red} and \textcolor[HTML]{315AD5}{blue}. The actions are marked {\color{pink}pink} and \textcolor[HTML]{71FFFF}{cyan}. The green checkmarks indicate correct name/action predictions, while the red crosses indicate incorrect ones.}
	\label{fig:figure1}
    \vspace{-1.5em}
\end{figure}

The traditional video captioning methods~\cite{lin2022swinbert, tang2021clip4caption, ye2022hierarchical, shen2023accurate, wang2024omnivid} achieve impressive results in open domains, which provide only a broad summary of the visual content, as shown in \cref{fig:figure1} (a). However, in some specific scenarios, such as sports live text broadcast, the players' identities and the actions are crucial, as audiences are particularly interested in these details. Such requirements cannot be satisfied by the general descriptions. Anonymous methods replace entities in descriptions with special tokens (\eg, [PLAYER] and [TEAM])~\cite{park2020identity, rao2024matchtime, mkhallati2023soccernet, rao2025towards} or numeric IDs (e.g., ``PLAYER0961226'')~\cite{wu2022sports}. They can provide detailed actions and distinguish different entity types. However, their descriptions are still missing entity identities, specifically player names, as shown in \cref{fig:figure1} (b).

Recently, some methods~\cite{qi2023goal, xi2024knowledge, xi2025eika} have introduced extra information, such as game news, player statistics, and the candidate player list, to generate descriptions with player identities. Due to the fact that the extra information is independent of the video clips, player identities in the descriptions are derived from the static extra information rather than from the video clips themselves. These methods do not truly recognize player identities from the video clips, but rather make inferences based on extra information. The player identities may be incorrect, as shown in~\cref{fig:figure1} (c).

The above methods fail to correctly describe players' identities, which has become a major backlog for sports video captioning. In fact, player identities are implied within the video clips, but most datasets~\cite{chen2011collecting, xu2016msr, caba2015activitynet, das2013thousand, qi2019sports, rao2024matchtime, zhang2024descriptive, rao2025towards} do not include the entity names, which makes it impossible to generate the identity-aware descriptions. Although some existing sports video captioning datasets include player names~\cite{qi2023goal, xi2024knowledge, xi2025eika}, they lack the necessary annotations to facilitate the development of models with visual mining capabilities. For instance, player bounding boxes could help build models to directly recognize player identities in visual content. If we could re-organize video clips around individual players to obtain the player-centric video clip set, it is possible to develop a player identification network for identity-aware video captioning.

In this paper, we collect 40 NBA full games during 2022-2023, and organize a player-centric video clip set to link each player with all of his video clips. Based on this set, a player identification network (PIN) is developed to extract players' visual features and names from the given videos. A bidirectional semantic interaction module (BSIM) is designed to augment the player visual features by interacting with the video features. The above two modules form an identity-related information extraction module (IRIEM), which outputs the identity-related multimodal embeddings. A learnable approach is adapted by utilizing a designed visual context learning module (VCLM) to learn video context information, which is then concatenated with the identity-related multimodal embeddings as the prompt for the LLM to generate descriptions with player names. To support this, we propose an identity-aware basketball video captioning dataset NBA-Identity, which involves 9,726 videos, 321 players with bounding boxes, and 9 major kinds of events.  

The main contributions of this paper are:
\begin{itemize}
    \item We propose a player-centric multimodal prompt generation network for LLM-based identity-aware basketball video captioning (LLM-IAVC),  which jointly utilizes player identity features and contextual video information to guide LLM in generating identity-aware descriptions.
    \item A large identity-aware basketball video captioning dataset NBA-Identity is constructed. This dataset contains the largest number of videos among identity-aware sports video captioning datasets. Its captions cover 9 major events in the basketball domain.
    \item Extensive experiments on our NBA-Identity and VC-NBA-2022~\cite{xi2024knowledge} datasets demonstrate that the proposed model achieves state-of-the-art performance without the need for additional knowledge assistance.
\end{itemize}

\section{Related Works}
\label{sec:related works}

\noindent \textbf{Video Captioning.} Video captioning aims to comprehend the video content and produce descriptions in natural language automatically. 
Recent researches~\cite{lin2022swinbert,  tang2021clip4caption, ye2022hierarchical, shen2023accurate, wang2024omnivid} primarily emphasize generation methods based on sequence learning, where a visual encoder extracts relevant visual features from videos, and a decoder then generates descriptions in a sequential manner. Researchers have attempted to employ various visual encoders, including ResNet~\cite{he2016deep}, Vision Transformer (ViT)~\cite{dosovitskiy2020image}, and C3D~\cite{hara2018can}, to extract different 2D/3D video features. In this work, our LLM-IAVC employs the encoder-decoder framework and integrates the pre-trained model Timesformer~\cite{bertasius2021space} as the visual encoder to encode the spatiotemporal information in videos. Unlike most models that use LSTM \cite{cho2014learning} or Transformer~\cite{vaswani2017attention} as the text decoder, we employ the LLM for the naturalness and accuracy of generated text. 

\noindent \textbf{Sports Video Captioning.} It aims to recognize player identities and key event types, generating accurate descriptions in fluent and natural language to ensure alignment with the video content. Anonymous methods utilize special tokens (\eg, [PLAYER] and [TEAM])~\cite{park2020identity, mkhallati2023soccernet, rao2024matchtime, rao2025towards} or entity types (\eg, ``Batter'' in baseball)~\cite{jhamtani2018learning, kim2020automatic} in descriptions to represent entities. These methods still cannot generate descriptions with specific entity names. Extra information based methods~\cite{qi2023goal, xi2024knowledge, xi2025eika} generate descriptions that include player names by incorporating extra game-related information. This type of method guesses player identities from a pool of candidate information but does not genuinely recognize players' identities from a visual perspective. Different from the above methods, this paper focuses on extracting identity-related information directly from the video clips, which is helpful for preserving the consistency of the descriptions and the video clips. 

\begin{figure*}[t!]
	\centering
	\includegraphics[width=16.5cm,height=8.0cm]{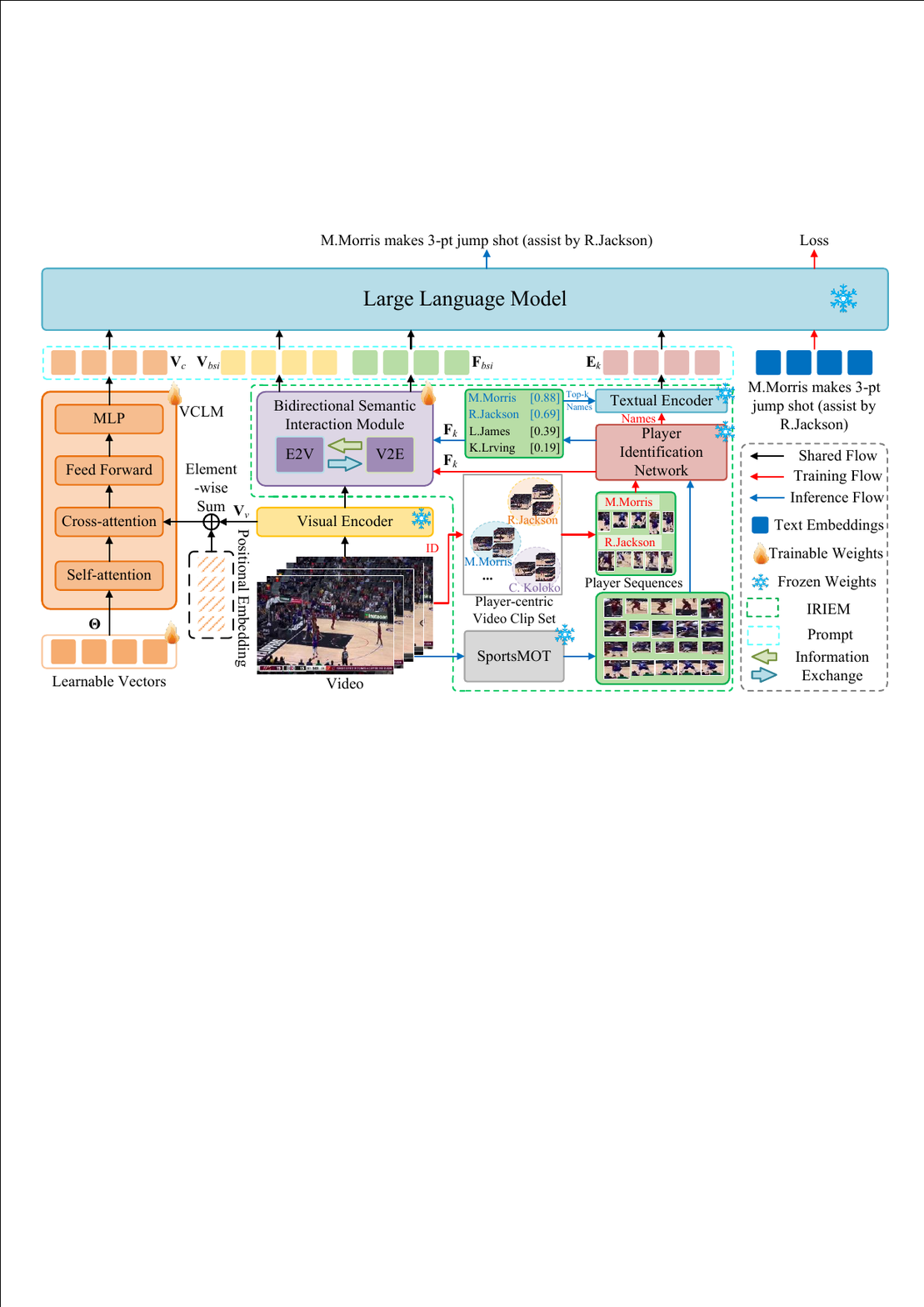}
	\caption{Illustration of the proposed identity-aware basketball video captioning model. During training, the player sequences corresponding to the input video are located from the player-centric video clip set based on the video ID, and their visual features are extracted via a pre-trained player identification network. During inference, a multi-object tracker extracts player sequences, and the player identification network extracts their visual features and identities. The top-$k$ players with the highest confidence are selected for further processing.}
	\label{fig:figure2}
\end{figure*}

\noindent \textbf{Large Language Models.} LLMs (\eg, LLaMa2~\cite{touvron2023llama} and GPT4~\cite{achiam2023gpt}) exhibit proficiency in a variety of language tasks given their powerful reasoning ability. Recent works~\cite{ozdemir2024enhancing, ren2024timechat, zanella2024harnessing} attempt to empower LLMs with visual understanding ability on downstream tasks like visual question-answering, video grounding, and video anomaly detection. With its impressive text generation capabilities, we employ various lightweight LLMs as the decoders, including GPT-2~\cite{radford2019language}, Qwen2.5~\cite{qwen2.5}, and Llama3.2~\cite{llama3.2}, to generate high-quality descriptions for the sports video captioning task. Instead of manually designing a sentence as a prompt, we automatically generate a multimodal prompt that includes identity-related information and visual context, helping the LLM better understand the video content.

\section{Methods}

As shown in \cref{fig:figure2}, LLM-IAVC mainly consists of three main components: 1) identity-related information extraction module (IRIEM), 2) visual context learning module (VCLM), and 3) LLM-based decoder. The IRIEM includes the player identification network (PIN) and the bidirectional semantic interaction module (BSIM). The red line represents the training flow and the blue line represents the inference flow. The black flow represents the shared process for training and inference. In the following subsections, we will describe each component in detail.

\subsection{Player Identification Network}

\begin{figure}[t!]
\centering
\includegraphics[width=8.2cm,height=5.3cm]{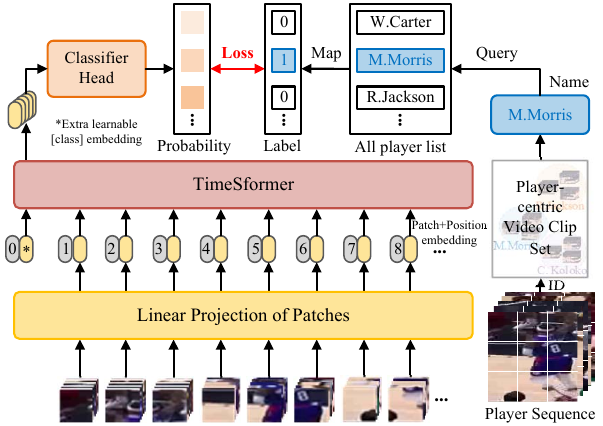}
	\caption{Training for the player identification network, including TimeSformer as the
visual backbone and a classification head.}
	\label{fig:figure3}
  \vspace{-1.0em}
\end{figure}

Since player identities are inherently embedded in video clips, a straightforward approach is to develop a player identification module that directly recognizes player identities from visual content. Based on the player names in captions, each video clip is associated with specific players. Therefore, all training video clips from 40 full games are automatically re-organized as the player-centric video clip set. The player sequences could be obtained by their bounding boxes. The player identification network is developed to extract the identity-related visual features from the player sequences. \cref{fig:figure3} shows the training process of the player identification network, which consists of TimeSformer as the visual backbone and a fully connected layer as the classification head. 

Let $P_{B}=\left \{ p_{1}, p_{2},\dots ,p_{n} \right \}$ represent the player sequence, where $p_{n}$ denotes the $n$-th the cropped bounding box. The player sequence feature $\mathbf{F}_{P} \in \mathbb{R}^{1 \times D_{time}}$ is extracted by TimeSformer~\cite{bertasius2021space}. Then  $\mathbf{F}_{P}$ is fed into a classification network to obtain classification probabilities.
\begin{equation}
o _{P} =\mathbf{W}_{f} \left ( \mathbf{F}_{P}  \right ) =\mathbf{W}_{f} \left ( \mathrm{M_{time}}\left (  P_{B} \right )   \right ),  
\label{eq:player_probability}
\end{equation}
where $\mathbf{W}_{f} \in \mathbb{R}^{D_{time} \times D_{c}}$ is a fully connected layer. $D_{time}$ is the hidden size of TimeSformer $\mathrm{M_{time}}$. $D_{c}$ is the number of player classes. The player identification network is trained with a cross-entropy loss:
\begin{equation}
\mathcal{L}_{player}=-\frac{1}{N_{b}} \sum_{n=1}^{N_{b}}\sum_{i=1}^{D_{C}}y_{n,i}\cdot \mathrm{log}\left ( p_{n,i} \right ),
\label{eq:player_loss}
\end{equation}
where $N_{b}$ is the batch size. $y_{n,i}$ and $p_{n,i}$ are the true label and predicted probability for sample $n$ and class $i$.

\subsection{Bidirectional Semantic Interaction Module}

The BSIM is designed to strengthen the association between player identity-related visual features and overall video content through information exchange, ensuring that player identities are more effectively integrated into the visual context. BSIM involves a multi-head self-attention module, a multi-head cross-attention module, and an MLP layer. The BSIM aims to associate video and player features and we define this operation for the multi-head self-attention mechanism with the input vector $\mathbf{I}$:
\begin{equation}
\mathrm{MSA}\left ( \mathbf{I} \right ) =\delta \left ( \frac{\mathbf{I}\mathbf{W}_{Q}\cdot \left ( \mathbf{IW}_{K} \right )^{\top} }{\sqrt{D} } \right )\cdot \mathbf{IW}_{V}, 
\label{eq:equation3}
\end{equation}
where $\mathbf{W}_{Q}$, $\mathbf{W}_{K}$ and $\mathbf{W}_{V}$ are learnable query, key, and value parameter matrices, respectively. $D$ is the dimension of the input vector $\mathbf{I}$. $\delta \left ( \cdot  \right )$ is the softmax function. The multi-head cross-attention mechanism is the variant of the multi-head self-attention mechanism, where the query comes from one input vector $\mathbf{I}_{1}$ while the key and value come from another input vector $\mathbf{I}_{2}$:
\begin{equation}
\mathrm{MCA}\left ( \mathbf{I}_{1}, \mathbf{I}_{2} \right ) =\delta \left ( \frac{\mathbf{I}_{1}\mathbf{W}_{Q}\cdot \left ( \mathbf{I}_{2}\mathbf{W}_{K} \right )^\mathrm {\top} }{\sqrt{D} } \right )\cdot \mathbf{I}_{2}\mathbf{W}_{V}.
\label{eq:equation4}
\end{equation}

\begin{figure}[t!]
\centering
\includegraphics[width=8.2cm,height=5.8cm]{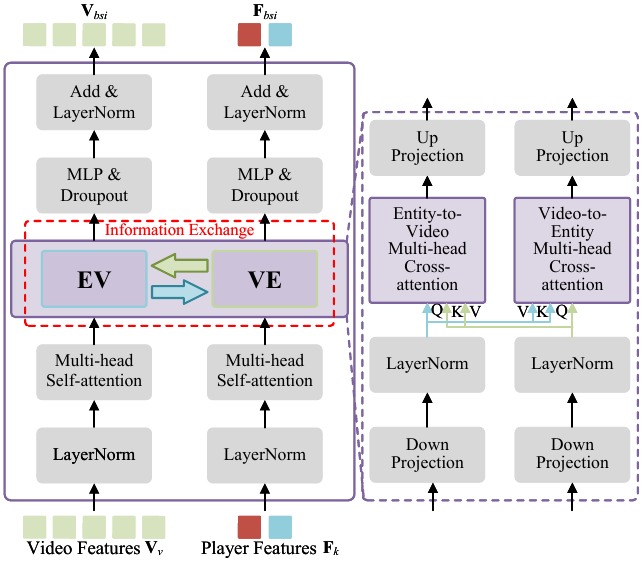}
    \caption{The bidirectional semantic interaction module.}
	\label{fig:figure4}
\end{figure}

The BSIM is shown in \cref{fig:figure4}. The visual features of the $k$ player sequences are concatenated into a player feature $\mathbf{F}_{k}\in \mathbb{R} ^{ k \times D_{time}}$, which is then interacted with the video feature $\mathbf{V}_{v}$ through the bidirectional semantic interaction module. Specifically, the video feature $\mathbf{V}_{v}$ and player feature $\mathbf{F}_{k}$ each pass through a multi-head self-attention layer for self-enhancement:
\begin{equation}
\begin{cases}
\mathbf{V}_{v}^{'} = \mathrm{MSA}_{1}\left ( \ell_{1} \left ( \mathbf{V}_{v} \right )  \right ) 
\\\mathbf{F}_{k}^{'} = \mathrm{MSA}_{2}\left ( \ell_{2} \left ( \mathbf{F}_{k} \right )  \right )

\end{cases},
\label{eq:equation5}
\end{equation}
where $\ell \left ( \cdot  \right )$ denotes the Layer Normalization layer.

Then $\mathbf{V}_{v}^{'}$ and $\mathbf{F}_{k}^{'}$ are associated through an information exchange sub-module:
\begin{equation}
\begin{cases}
\mathbf{V}_{v}^{''} =  \mathbf{W}_{u1}\mathrm{MCA}_{ev}\left ( \ell_{3}\left ( \mathbf{W}_{d1}\mathbf{V}_{v}^{'} \right ), \ell_{4}\left ( \mathbf{W}_{d2}\mathbf{F}_{k}^{'} \right ) \right ) 
 
\\\mathbf{F}_{k}^{''} =  \mathbf{W}_{u2}\mathrm{MCA}_{ve}\left ( \ell_{4}\left ( \mathbf{W}_{d2}\mathbf{F}_{k}^{'} \right ), \ell_{3}\left ( \mathbf{W}_{d1}\mathbf{V}_{v}^{'} \right ) \right ) 
\end{cases},
\label{eq:equation6}
\end{equation}
where $\mathbf{W}_{d}\in \mathbb{R} ^{ D_{time} \times D_{down}}$ and $\mathbf{W}_{u}\in \mathbb{R} ^{ D_{down} \times D_{time}}$ denote the linear down- and up-projection matrices. $\mathrm{MCA}_{ev}  \left ( \cdot  \right )$ and $\mathrm{MCA}_{ve}  \left ( \cdot  \right )$ denote entity-to-video multi-head cross-attention and video-to-entity multi-head cross-attention, respectively.

Finally, the outputs of BSIM are as follows:
\begin{equation}
\begin{cases}
\mathbf{V}_{bsi} = \ell_{5} \left ( \varrho_{1} \left ( \mathrm{MLP}_{1}\left ( \phi_{1} \left ( \mathbf{V}_{v}^{''} \right ) \right ) \right )   + \mathbf{V}_{v}^{''}\right )  
 
\\\mathbf{F}_{bsi} =  \ell_{6} \left ( \varrho_{2} \left ( \mathrm{MLP}_{2}\left ( \phi_{2} \left ( \mathbf{F}_{k}^{''} \right ) \right ) \right )   + \mathbf{F}_{k}^{''} \right ) 
\end{cases},
\label{eq:equation7}
\end{equation}
where $\varrho \left ( \cdot  \right )$ demotes the Dropout layer. $\phi \left ( \cdot  \right )$ denotes the GELU activation function~\cite{hendrycks2016gaussian}. And $\mathrm{MLP}\left ( \cdot  \right ) $ denotes the multilayer perceptron layer.

\subsection{Visual Context Learning Module}
\vspace{-5pt}
The visual context learning module (VCLM) aims to learn the video context information $\mathbf{V}_{c}\in \mathbb{R} ^{ N_{q} \times D_{time}}$. As shown in \cref{fig:figure2}, we randomly initialize $N_{q}$ learnable query vectors $\mathbf{\Theta}\in \mathbb{R} ^{ N_{q} \times D_{time}}$. VCLM extracts a kind of ``global" and ``overview" semantic information from the original video features through learnable query vectors to help the model capture the overall dynamic background of the video content. Learnable vectors not only provide contextual information, but also act as a bridge between the visual and textual spaces. These vectors can be automatically adjusted during training to better fit the semantic mapping between vision and text. We use self- and cross-attention  to aggregate video context information:
\begin{equation}
\mathbf{V}_{self} =\delta \left ( \frac{\mathbf{\Theta} \mathbf{W}^{Q_{1}}\cdot \left ( \mathbf{\Theta} \mathbf{W}^{K_{1}}  \right )^{\mathrm {\top}} }{\sqrt{D_{time} } } \right )\cdot \mathbf{\Theta} \mathbf{W}^{V_{1}},
\label{eq:equation8}
\end{equation}
\begin{equation}
\mathbf{V}_{pv} =\tau_{p} + \mathbf{V}_{v},
\label{eq:equation9}
\end{equation}
\begin{equation}
\mathbf{V}_{cross} =\delta \left ( \frac{\mathbf{V}_{self} \mathbf{W}^{Q_{2}}\cdot \left ( \mathbf{V}_{pv}\mathbf{W}^{K_{2}} \right )^{\mathrm {\top}}}{\sqrt{D_{time}} }  \right ) \cdot \mathbf{V}_{pv}\mathbf{W}^{V_{2}}  ,
\label{eq:equation10}
\end{equation}

\begin{equation}
\mathbf{V}_{c} =\mathrm{FFN}\left ( \mathrm{MLP}\left ( \mathbf{V}_{cross} \right )  \right ) ,
\label{eq:equation11}
\end{equation}
where $\mathbf{V}_{v}\in\mathbb{R} ^{ N_{v}  \times D_{time}}$ is the video feature extracted by TimeSformer~\cite{bertasius2021space}. $N_{v}$ is the number of frames. $\tau$ is the position embedding and helps the model distinguish the sequence of frames and capture temporal dependencies.  $\mathbf{W}^{{Q}_{1}}\in \mathbb{R} ^{ D_{time}  \times D_{time}}$, $\mathbf{W}^{{K}_{1}}\in \mathbb{R} ^{ D_{time}  \times D_{time}}$, $\mathbf{W}^{{V}_{1}}\in \mathbb{R} ^{ D_{time} \times D_{time}}$, $\mathbf{W}^{{Q}_{2}}\in \mathbb{R} ^{ D_{time}  \times D_{time}}$, $\mathbf{W}^{{K}_{2}}\in \mathbb{R} ^{ D_{time}  \times D_{time}}$ and $\mathbf{W}^{{V}_{2}}\in \mathbb{R} ^{ D_{time}  \times D_{time}}$ are learnable matrices. $\mathrm{FFN}\left ( \cdot  \right ) $ denotes the feed-forward network.

\subsection{Large Language Model Based Decoder}

We use a pre-trained LLM as the decoder. First, the pre-trained LLM is employed to extract the corresponding textual feature $\mathbf{E}_{n}\in \mathbb{R} ^{ 1 \times D_{llm}}$ for each player’s name in the player sequence. Then, we concatenate $k$ name features into $\mathbf{E}_{k}\in \mathbb{R} ^{ k \times D_{llm}}$. To enhance generalization and adaptability, a multimodal prompt is provided to the LLM, which includes video context information $\mathbf{V}_{c}$, video feature $\mathbf{V}_{bsi}$, player feature $\mathbf{F}_{bsi}$, and player name feature $\mathbf{E}_{k}$. This prompt enables the model to generate captions $\mathbf{\hat{C}}$ that accurately incorporate player identities.
\begin{equation}
\mathbf{\hat{C}} =\Psi_{llm}\left (\left [ \mathbf{W}_{c}\mathbf{V}_{c}, \mathbf{W}_{v}\mathbf{V}_{bsi}, \mathbf{W}_{f}\mathbf{F}_{bsi}, \mathbf{W}_{n}\mathbf{E}_{k}  \right ]  \right ) ,
\label{eq:equation12}
\end{equation}
where $\mathbf{W}_{c}\in \mathbb{R} ^{ D_{time} \times D_{llm}}$, $\mathbf{W}_{v}\in \mathbb{R} ^{ D_{time} \times D_{llm}}$, $\mathbf{W}_{f}\in \mathbb{R} ^{ D_{time} \times D_{llm}}$ and $\mathbf{W}_{n}\in \mathbb{R} ^{ D_{llm} \times D_{llm}}$ are linear projection layers that map each feature into the embedding space of the LLM. $D_{llm}$ is the hidden size of the LLM.  $\Psi_{llm} \left ( \cdot  \right )$ denotes the LLM-based decoder, and $\left [ \cdot  \right ] $ denotes the concatenate function in Python.

\subsection{Training and 
Inference}

During the training stage, the video\_id and game\_id are utilized to query the player-centric video clip set, retrieving the corresponding key player sequences and player names. The player identification network extracts player visual features from player sequences, while player names are encoded into textual features using an LLM. Following the conventional training pattern for caption generation, we utilize cross-entropy loss to optimize our LLM-IAVC:
\begin{equation}
\mathcal{L}_{\Phi }  =-\sum_{t=1}^{N_{g}} \mathrm{log}  \left ( \mathrm{P}  \left ( \beta_{t}^{*}  | \beta_{0:t-1}^{*}, \mathbf{F}_{prompt}; \Phi  \right )  \right ),
\label{eq:equation13}
\end{equation}

\begin{equation}
\mathbf{F}_{prompt} = \left [ \mathbf{W}_{c}\mathbf{V}_{c}, \mathbf{W}_{v}\mathbf{V}_{bsi}, \mathbf{W}_{f}\mathbf{F}_{bsi}, \mathbf{W}_{n}\mathbf{E}_{k}  \right ],
\label{eq:equation14}
\end{equation}
where $\left \{  \beta_{0}^{*}, \beta_{1}^{*}, ..., \beta_{N_{g}}^{*}\right \}$ is the set of ground-truth tokenized tokens. $N_{g}$ denotes the number of tokens. And $\Phi$ denotes the optimized parameters.

At the inference stage, there are some differences compared to the training stage. We utilize the pre-trained multi-object tracking model SportsMOT~\cite{cui2023sportsmot} to extract multiple player sequences. These sequences are then processed by the player identification network to recognize player identities. The identified sequences are ranked based on confidence scores, and the top-$k$ sequences are selected. The visual features of these top-$k$ players are concatenated to form the player feature representation $\mathbf{F}_{k}$. Additionally, the names of the top-$k$ players are encoded into textual features using a pre-trained LLM, resulting in a concatenated feature representation $\mathbf{E}_{k}$. Finally, as in \cref{eq:equation12}, this concatenated feature serves as input prompt to the LLM, generating the corresponding description.

\section{Experiments}
\label{sec:experiment}
LLM-IAVC is compared with existing video captioning models on NBA-Identity and VC-NBA-2022~\cite{xi2024knowledge} datasets. We further conduct ablation studies to assess the contribution of each component in LLM-IAVC.

\begin{table}[tb!]
\resizebox{\columnwidth}{!}{%
\begin{tabular}{lccccccc}
\hline
\textbf{Dataset}        &  \textbf{Domain} & \textbf{Bbox} & \textbf{Video} & \textbf{Sentence} & \textbf{One} & \textbf{Multi.} & \textbf{Total Dur(h)} \\ \hline
\textbf{Anonymous}      &                     &               &                &                   &              &                 &                       \\
MSVD~\cite{chen2011collecting}                    & Open            & No            & 1.9k           & 70k               &              & \checkmark               & 5.3                   \\
ActivityNet-Caption~\cite{caba2015activitynet}     & Open            & No            & 20k            & 100k              &              & \checkmark               & 840                   \\
MSR-VTT~\cite{xu2016msr}                 & Open            & No            & 10k            & 200k              &              & \checkmark               & 41.2                  \\
TACoS~\cite{regneri2013grounding}                   & Cooking         & No            & 127            & 11.8              &              & \checkmark               & -                     \\
YouCook~\cite{das2013thousand}                 & Cooking         & No            & 88             & 2.7k              &              & \checkmark               & 2.3                   \\
YouCook2~\cite{zhou2018towards}                & Cooking         & No            & 2k             & 15.4              &              & \checkmark               & 176                   \\
SVCDV~\cite{qi2019sports}                   & Volleyball      & No            & 4.8k           & 44k               & \checkmark            &                 & -                     \\
Soccer captioning~\cite{hammoudeh2022soccer}       & Soccer          & No            & 22k           & 22k               & \checkmark            &                 & -                 \\
SoccerNet-Caption~\cite{mkhallati2023soccernet}       & Soccer          & No            & 0.9k           & 36k               & \checkmark            &                 & 715.9                 \\
MatchTime~\cite{rao2024matchtime}               & Soccer          & No            & 0.9k           & 36k               & \checkmark            &                 & 715.9                 \\
SoccerReplay-1988~\cite{rao2025towards}       & Soccer          & No            & 150k           & 150k              & \checkmark            &                 & 3,323.0                \\
FSN~\cite{yu2018fine}                     & Basketball      & No            & 2k             & 6.5k              & \checkmark            &                 & -                     \\
BH-Commentary~\cite{zhang2024descriptive}           & Basketball      & No            & 4.3k           & 4.3k              & \checkmark            &                 & 10.1                  \\ 
NSVA~\cite{wu2022sports}           & Basketball      & Yes            & 32k           & 44.6k              & \checkmark            &                 & 84.8                  \\ \hline
\textbf{Identity-aware} &                          &               &                &                   &              &                 &                       \\
MPII-MD~\cite{rohrbach2015dataset}                 & Movie           & No            & 68.3k          & 68.3k             & \checkmark            &                 & 73                    \\
Movie101~\cite{yue2023movie101}                & Movie           & No            & 30k            & 30k               & \checkmark            &                 & 92                    \\
Goal~\cite{qi2023goal}                    & Soccer          & No            & 8.9k           & 22k               & \checkmark            &                 & 25.5                  \\
VC-NBA-2022~\cite{xi2024knowledge}             & Basketball      & No            & 3.9k           & 3.9k              & \checkmark            &                 & -                     \\
\rowcolor[HTML]{DDDDDD} 
NBA-Identity            & Basketball      & Yes           & 9.7k           & 9.7k              & \checkmark            &                 & 8.9                   \\ \hline
\end{tabular}
}
\caption{Comparison of existing video captioning datasets. ``Bbox'' denotes whether the dataset includes entity bounding boxes. ``One'' means each video is paired with only one caption, while ``Multi.'' means multiple different captions.}
\label{tab:example1}
\vspace{-1.0em}
\end{table}

\subsection{Datasets and Metrics}
We collect 40 games data from the NBA 2022-2023 season to construct an identity-aware basketball video caption dataset \textbf{NBA-Identity}. It includes 9,726 videos, 321 players with bounding boxes, and 9 major kinds of events (\eg, ``block'', ``foul'', ``defensive rebound'', ``offensive rebound'', ``turnover'', ``two-point (2-pt) shot'', ``three-point (3-pt) shot'', ``layup'' and ``assist''). Notably, ``foul'' includes ``personal foul'', ``personal take foul'', ``shooting foul'', ``offensive foul'', and ``loose ball foul''. Similarly, ``turnover'' includes ``lost ball; steal by [player]'', ``bad pass; steal by [player]'', ``lost ball'', ``bad pass'', ``offensive foul'', ``traveling'', ``step out of bounds'', ``discontinued dribble'', and ``out of bounds lost ball''. Our NBA-Identity covers 26 fine-grained event types. And it is divided into training and testing sets based on game instances. 8,667 clips from 35 games for training, 1,059 clips from 5 games for testing. As shown in~\cref{tab:example1}, open-domain datasets (including cooking) typically provide multiple captions per video clip, whereas sports and movie datasets usually have only one. This may be due to: 1) the higher difficulty and cost of obtaining domain-specific annotations, and 2) the uniqueness of terminology in specialized fields like sports. To facilitate identification, we also provide key player bounding boxes.

\textbf{VC-NBA-2022} is an identity-aware basketball video captioning dataset, which includes 9 types of fine-grained shooting events, 286 players, and over 3.9k videos. The training set contains 3,162 videos and the testing set contains 786 videos. We employ the typical captioning metrics, including BLEU~\cite{papineni2002bleu}, Rouge-L~\cite{lin2004rouge}, METEOR~\cite{banerjee2005meteor} and CIDEr~\cite{vedantam2015cider} to evaluate the performance of LLM-IAVC. 

Compared to VC-NBA-2022, Our NBA-Identity extends VC-NBA-2022 with the following key differences: 1) More video clips; 2) More event types; 3) More realistic video clips: Video clips of NBA-Identity are in MP4 format with variable lengths based on event duration, while clips in VC-NBA-2022 are uniformly sampled to 72 frames. 4) All captions in NBA-Identity are directly obtained from live broadcast platforms without any modifications, making them authentic, while the textual annotations in VC-NBA-2022 have been manually revised into structured subject-verb-object formats. 5) Captions include shot distance, making generation more challenging and realistic. \textbf{More details about the proposed dataset and metrics can be found in Supplementary materials}.

\subsection{Implementation Details}

The hidden size $D_{time}$ of TimeSformer is 768. Each video clip in NBA-Identity has up to 60 frames, with frame and player image sizes of $224 \times 224$. Since descriptions of the dataset mention at most two players, the top-2 player sequences are selected for visual feature extraction. We stack 1 visual context learning module block and 1 bidirectional semantic interaction module layer for LLM-IAVC. The number of learnable query vectors is set to 32 for visual context learning module. And the $D_{down}$ is set to 512 for bidirectional semantic interaction module. For player identification network training, the Adam~\cite{kingma2014adam} optimizer is used with a learning rate of 5e-5 for 50 epochs, covering 321 player identities. The number of images in player sequence is set to 20. LLM-IAVC is trained for 100 epochs on both datasets. Variants of LLM-IAVC utilize different LLMs as decoders, including GPT-2, Qwen2.5, and Llama3.2, with training parameters detailed in~\cref{tab:example2}. The Adam optimizer is used across all variants, with a beam search size of 5 for decoding. All experiments run on an NVIDIA 4090 GPU. For example, with LLama3.2-3B (batch = 8), training and testing per epoch take 221.2s and 115.3s, respectively, with 0.3s per caption generation.

\begin{table}[t!]
\footnotesize
\centering
\begin{tabular}{@{}lccc@{}}
\toprule
\textbf{Model}        & \textbf{Learning-rate} & \textbf{Hidden size} & \textbf{Batch size} \\ \midrule
GPT-2~\cite{radford2019language}        & 5e-5          & 768         & 64         \\
Qwen2.5-0.5B~\cite{qwen2.5} & 7e-5          & 896         & 32         \\
Qwen2.5-1.5B~\cite{qwen2.5} & 5e-5          & 1,536        & 32         \\
Qwen2.5-3B~\cite{qwen2.5}   & 1e-5          & 2,048        & 8          \\
Llama3.2-1B~\cite{llama3.2}  & 5e-5          & 2,048        & 32         \\
Llama3.2-3B~\cite{llama3.2}  & 7e-6          & 3,072        & 8          \\ \bottomrule
\end{tabular}%
\vspace{-2pt}
\caption{Training parameters for different variants of LLM-IAVC.}
\label{tab:example2}
\vspace{-1.5em}
\end{table}

\begin{table*}[t!]
\centering
\resizebox{0.9\textwidth}{!}{%
\begin{tabular}{@{}lcccccccc@{}}
\toprule
\multicolumn{1}{l|}{\multirow{2}{*}{\textbf{Method}}} & \multicolumn{4}{c|}{\textbf{NBA-Identity}}                                                                                  & \multicolumn{4}{c}{\textbf{VC-NBA-2022}}                                                                                   \\ \cmidrule(l){2-9} 
\multicolumn{1}{l|}{} & \multicolumn{1}{c}{\textbf{CIDEr}} & \multicolumn{1}{c}{\textbf{METEOR}} & \multicolumn{1}{c}{\textbf{Rouge-L}} & \multicolumn{1}{c|}{\textbf{BLEU-4}} & \multicolumn{1}{c}{\textbf{CIDEr}} & \multicolumn{1}{c}{\textbf{METEOR}} & \multicolumn{1}{c}{\textbf{Rouge-L}} & \multicolumn{1}{c}{\textbf{BLEU-4}} \\ \midrule
\multicolumn{9}{c}{\cellcolor[HTML]{DDDDDD}\textbf{Traditional methods}}
\\ \midrule
\multicolumn{1}{l|}{Clip4Caption~\cite{tang2021clip4caption}}           & 21.7                      & 21.0                       & 37.6                        & \multicolumn{1}{c|}{6.5}    & 70.4                      & 26.7                       & 51.2                        & 28.8                       \\
\multicolumn{1}{l|}{SwinBERT~\cite{lin2022swinbert}}                & 23.5                      & 20.7                       & 37.1                        & \multicolumn{1}{c|}{7.5}    & 69.1                      & 26.5                       & 49.0                        & 28.4                       \\
\multicolumn{1}{l|}{CoCap~\cite{shen2023accurate}}                   & 22.7                      & 20.0                       & 37.7                        & \multicolumn{1}{c|}{7.5}    & 70.5                      & 27.4                       & 50.7                        & 28.9                       \\
\multicolumn{1}{l|}{OmniViD~\cite{wang2024omnivid}}                 & 39.6                      & 21.6                       & 37.7                        & \multicolumn{1}{c|}{8.3}    & 71.2                      & 27.5                       & 50.7                        & 29.2                       \\ \midrule
\multicolumn{9}{c}{\cellcolor[HTML]{DDDDDD}\textbf{Extra inforamtion enhanced methods}}                                                                                                                                                                                                                                \\ \midrule
\multicolumn{1}{l|}{KEANet~\cite{xi2024knowledge}}                  & 50.1                      & 22.7                       & 38.1                        & \multicolumn{1}{c|}{8.5}    & 138.5                     & 28.0                       & 54.9                        & 32.4                       \\
\multicolumn{1}{l|}{EIKA~\cite{xi2025eika}}                    & 55.7                      & 22.7                       & 38.6                        & \multicolumn{1}{c|}{9.0}    & 140.7                     & 29.5                       & 56.8                        & 36.7                       \\ \midrule
\multicolumn{9}{c}{\cellcolor[HTML]{DDDDDD}\textbf{LLM-IAVC without prompt}}                                                                                                                                                                                                                                             \\ \midrule
\multicolumn{1}{l|}{LLM-IAVC (G)}          & 46.8                      & 17.5                       & 34.0                        & \multicolumn{1}{c|}{9.2}    & 87.4                      & 27.7                       & 53.8                        & 30.5                       \\
\multicolumn{1}{l|}{LLM-IAVC (Q-0.5B)}   & 54.3                      & 17.9                       & 33.8                        & \multicolumn{1}{c|}{11.5}   & 88.1                      & 27.6                       & 54.2                        & 30.8                       \\
\multicolumn{1}{l|}{LLM-IAVC (Q-1.5B)}   & 58.8                      & 18.8                       & 38.5                        & \multicolumn{1}{c|}{12.3}   & 90.0                      & 27.5                       & 55.2                        & 31.0                       \\
\multicolumn{1}{l|}{LLM-IAVC (Q-3B)}     & 61.8                      & 18.2                       & 39.0                        & \multicolumn{1}{c|}{14.7}   & 101.2                     & 28.0                       & 55.5                        & 32.2                       \\
\multicolumn{1}{l|}{LLM-IAVC (L-1B)}    & 60.3                      & 18.0                       & 33.2                        & \multicolumn{1}{c|}{14.0}   & 90.0                      & 27.3                       & 55.3                        & 31.4                       \\
\multicolumn{1}{l|}{LLM-IAVC (L-3B)}    & 66.5                      & 18.2                       & 34.5                        & \multicolumn{1}{c|}{14.2}   & 103.1                     & 28.5                       & 55.7                        & 32.8                       \\ \midrule
\multicolumn{9}{c}{\cellcolor[HTML]{DDDDDD}\textbf{LLM-IAVC with prompt}}                                                                                                                                                                                                                                                \\ \midrule
\multicolumn{1}{l|}{\textbf{LLM-IAVC (G)}}          & 94.1                      & 22.0                       & 35.4                        & \multicolumn{1}{c|}{17.0}   & 139.0                     & 29.0                       & \textcolor{blue}{\underline{57.8}}                        & 37.2                       \\
\multicolumn{1}{l|}{\textbf{LLM-IAVC (Q-0.5B)}}   & 87.6                      & 21.5                       & 36.4                        & \multicolumn{1}{c|}{16.0}   & 139.4                     & 29.2                       & 55.7                        & 36.8                       \\
\multicolumn{1}{l|}{\textbf{LLM-IAVC (Q-1.5B)}}   & 97.1                      & 22.1                       & 40.0                        & \multicolumn{1}{c|}{16.7}   & 141.5                     & 30.1                       & 57.4                        & 37.8                       \\
\multicolumn{1}{l|}{\textbf{LLM-IAVC (Q-3B)}}     & \textcolor{blue}{\underline{103.5}}                    & \textcolor{blue}{\underline{22.9}}                       & \textcolor{blue}{\underline{41.2}}                        & \multicolumn{1}{c|}{\textcolor{blue}{\underline{17.7}}}   & \textcolor{blue}{\underline{147.8}}                     & \textcolor{blue}{\underline{30.7}}                       & 56.8                        & \textcolor{blue}{\underline{38.3}}                       \\
\multicolumn{1}{l|}{\textbf{LLM-IAVC (L-1B)}}    & 88.3                      & 21.6                       & 40.1                        & \multicolumn{1}{c|}{15.3}   & 139.3                     & 30.0                       & 57.3                        & 37.7                       \\
\multicolumn{1}{l|}{\textbf{LLM-IAVC (L-3B)}}    & \textcolor{red}{\textbf{105.3}}                     & \textcolor{red}{\textbf{23.0}}                       & \textcolor{red}{\textbf{42.5}}                        & \multicolumn{1}{c|}{\textcolor{red}{\textbf{18.8}}}   & \textcolor{red}{\textbf{150.7}}                     & \textcolor{red}{\textbf{30.9}}                       & \textcolor{red}{\textbf{57.9}}                        & \textcolor{red}{\textbf{38.6}}                       \\ \bottomrule
\end{tabular}}
\caption{Quantitative comparison results on NBA-Identity and VC-NBA-2022 datasets. Within each unit, we denote the best performance in \textcolor{red}{\textbf{RED}} and the second-best performance in \textcolor{blue}{\underline{BLUE}}. G, Q, and L represent GPT-2, Qwen2.5, and Llama3.2, respectively.}
\label{tab:example3}
\end{table*}

\subsection{Results on NBA-Identity and VC-NBA-2022}

\noindent \textbf{Compared models.} The proposed LLM-IAVC is compared with 5 advanced video captioning models, including Clip4Caption~\cite{tang2021clip4caption}, SwinBERT~\cite{lin2022swinbert}, CoCap~\cite{shen2023accurate}, OmniViD~\cite{wang2024omnivid}, KEANet~\cite{xi2024knowledge} and EIKA~\cite{xi2025eika}. Clip4Caption employs CLIP~\cite{radford2021learning} to acquire the aligned visual-text representation for better generating the text descriptions. SwinBERT introduces Video Swin Transformer~\cite{liu2022video} to encode spatiotemporal representations from video frames. CoCap utilizes the motion vector feature, residual feature, and video feature through the action encoder. Then, the fused features are sent to the decoder to generate video captions. OmniViD is a unified generative framework that can address various video tasks, including action recognition, video captioning, video question answering, dense video captioning, and visual object tracking. KEANet and EIKA utilize the candidate player list to generate the descriptions with player identities.

\noindent \textbf{Quantitative results  analysis.} In \cref{tab:example3}, LLM-IAVC significantly outperforms the other 6 methods on CIDEr. This is due to CIDEr's sensitivity to unique or infrequent words like names, with accurate name prediction boosting scores. Traditional methods perform worse as they cannot directly recognize player identities from video. While KEANet and EIKA use a predefined candidate player list, they lack visual recognition, leading to less accurate player names. 

LLM-IAVC includes multiple variants with different LLM decoders, such as GPT-2~\cite{radford2019language}, Qwen 2.5~\cite{qwen2.5} and Llama 3.2~\cite{llama3.2}. Each variant takes either visual context alone or a multimodal prompt combining visual and identity features as input. Results show poor performance with visual context alone, while multimodal prompts significantly enhance generation quality. For example, adding prompts boosts the CIDEr score of LLM-IAVC (Llama 3.2-3B) by 40.9\% and BLEU-4 by 4.3\%, as the prompts provide richer visual context and player identity information for better video understanding. Among decoders, LLaMA outperforms Qwen and GPT-2 due to its superior architecture and pre-training process, leading to better contextual understanding and generation.

Due to resource limitations, we are unable to develop larger models. However, our approach is lightweight, allowing for easy deployment and training. These results underscore the capability of our model to generate accurate names and fine-grained actions in live text broadcast task.

\begin{figure*}[t!]
    \centering
    \includegraphics[width=16.5cm,height=8.0cm]{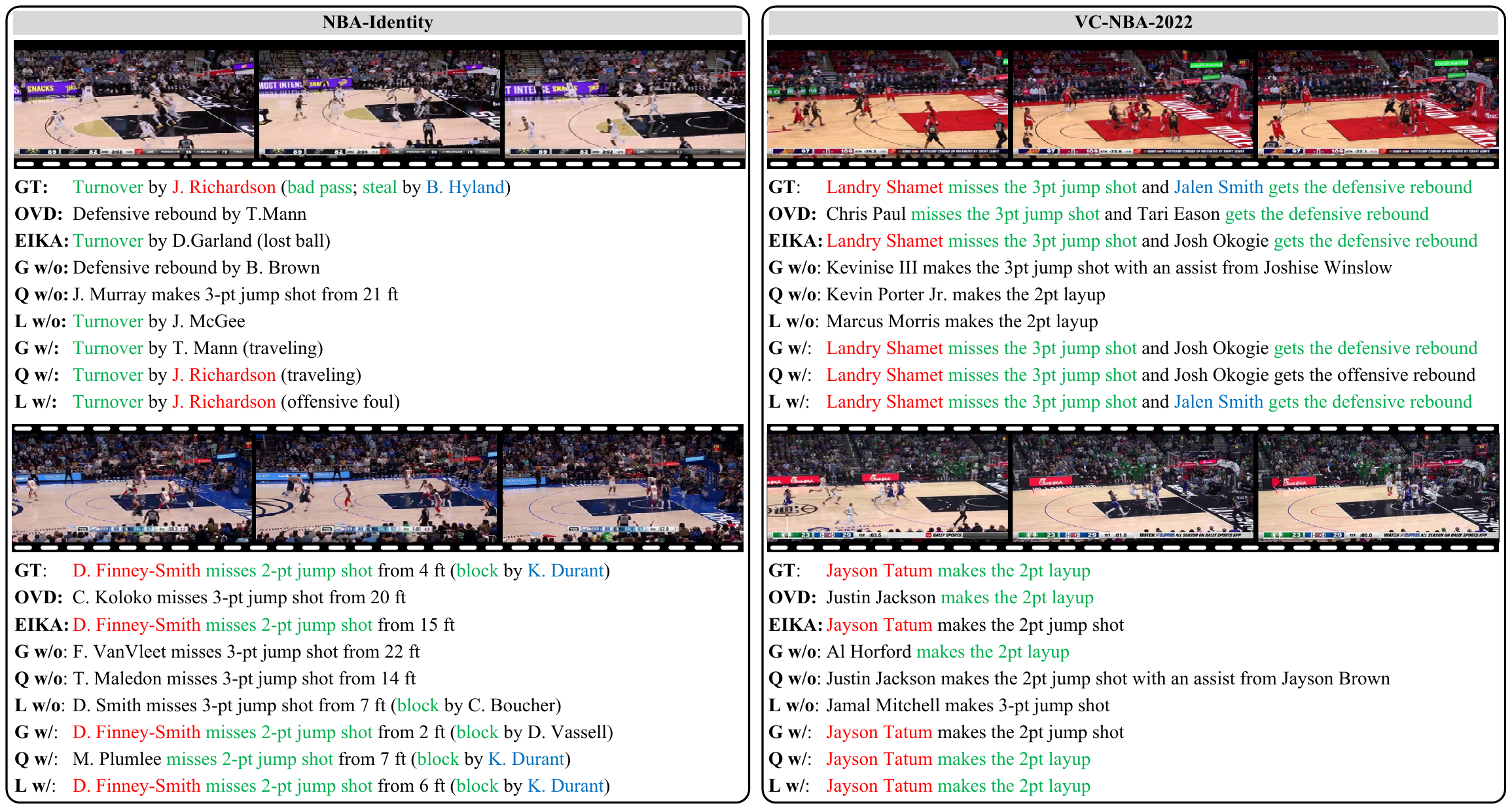}
    \caption{Qualitative results on NBA-Identity and VC-NBA-2022 datasets. ``Model w/o'' refers to the model using only video as input, without a prompt. ``Model w/'' refers to the model with a multimodal prompt as input. G, Q, and L denote GPT-2, Qwen2.5-3B, and Llama3.2-3B, respectively. Different entity names are marked {\color{red}red} and {\color{blue}blue}. And the specific visual scenes are marked {\color{green}green}.}
    \label{fig:figure5}
\end{figure*}

\noindent \textbf{Qualitative results analysis.}
\cref{fig:figure5} presents qualitative results on NBA-Identity and VC-NBA-2022. LLM-IAVC takes two input types: video-only and multimodal prompts. The video-only model struggles to understand player identities and actions, leading to inaccurate names, though they may resemble real names (\eg, ``Justin Jackson'' vs. ``Jayson Tatum''). By integrating video context and player identities, the generation quality is significantly improved, enabling more accurate descriptions of player names and actions. Moreover, shot distance is crucial in practical applications. It can be observed that traditional and extra information enhanced methods fail to accurately determine shooting distance. While our method doesn't explicitly predict it in NBA-Identity, providing richer and more detailed features as prompts allows the model to approximate the real shot distance more closely (\eg, ``6 ft'' vs. ``4 ft''). Though distance estimation remains a challenge, our model is a flexible and extensible baseline that can incorporate a distance evaluation module for improved accuracy.

\subsection{Ablation Study} 

\noindent \textbf{Backbone of player identification network.}
We compare different backbones for the player identification network in \cref{tab:example4}, treating player identification as a classification task where each class represents a player identity. We employ Multi-class Classification Accuracy (MCA) and Mean Per Class Accuracy (MPCA) metrics from the group activity recognition domain~\cite{wu2024learning} to assess the performance of the player identification network. A higher MCA indicates stronger overall recognition ability across all player categories. A higher MPCA reflects a better measure of the model’s generalization ability, especially in scenarios with imbalanced player categories. The TimeSformer-based~\cite{bertasius2021space} player identification network achieves the highest MCA (91.40\%) and MPCA (89.13\%), demonstrating superior accuracy and balanced player recognition.

\begin{table}[]
\footnotesize
\centering
\begin{tabular}{@{}l|c|cc@{}}
\toprule
\textbf{Backbone}          & \textbf{Input size} & \textbf{MCA}    & \textbf{MPCA}   \\ \midrule
ResNet-18~\cite{he2016deep}         & 320×180    & 86.46\% & 79.74\% \\
ResNet-50~\cite{he2016deep}         & 320×180    & \underline{90.12}\% & \underline{85.30}\% \\
C3D~\cite{hara2018can}               & 320×180    & 50.86\% & 43.32\% \\
I3D~\cite{carreira2017quo}               & 320×180    & 60.24\% & 57.20\% \\
Vison Transformer~\cite{dosovitskiy2020image} & 224×224    & 87.12\% & 80.06\% \\
\textbf{TimeSformer}~\cite{bertasius2021space}       & 224×224    & \textbf{91.40\%} & \textbf{89.13\%} \\ \bottomrule
\end{tabular}%
\caption{Impact of different backbones on player classification.}
\label{tab:example4}
\vspace{-0.5em}
\end{table}

\begin{table}[tb!]
\footnotesize
\centering
\begin{tabular}{@{}c|ccc|cc@{}}
\toprule
\textbf{Model} & \textbf{VCLM}         & \textbf{PIN}        & \textbf{BSIM}       & \textbf{CIDEr} & \textbf{BLEU-4} \\ \midrule
\ding{172}     &            &            &            & 66.5  & 14.2   \\
\ding{173}     & \checkmark &            &            & 66.8  & 15.0   \\
\ding{174}     &            & \checkmark &            & 95.1  & 16.4   \\
\ding{175}     & \checkmark & \checkmark &            & 99.7  & 17.7   \\
\ding{176}     &            & \checkmark & \checkmark & \underline{103.4} & \underline{18.1}   \\
\ding{177}     & \checkmark & \checkmark & \checkmark & \textbf{105.3} & \textbf{18.8}   \\ \bottomrule
\end{tabular}%
\caption{Ablation on NBA-Identity.}
\label{tab:example5}
\vspace{-2.0em}
\end{table}

\noindent \textbf{Contribution of each component.}
As shown in \cref{tab:example5}, ablation experiments assess the contributions of the visual context learning module (VCLM), player identification network (PIN), and bidirectional semantic interaction module (BSIM). Without any components, model~\ding{172} performs poorly due to the challenge of identifying players in complex sports scenes. Adding VCLM (model \ding{173}) slightly improves performance by helping the model focus on key video regions instead of scanning all frames. While contextual information enhances feature connections, the limited improvement reflects the difficulty of recognizing player identities from visual content.

\begin{table}[tb!]
    \footnotesize
    \centering
    \begin{tabular}{@{}c|cc|cc@{}}
        \toprule
        \textbf{Model} & \textbf{Video}     & \textbf{Player}  & \textbf{CIDEr} & \textbf{BLEU-4}  \\ \midrule
        \ding{172}     &  \checkmark          &             & 101.7  & 18.0   \\
        \ding{173}     &  &  \checkmark                     & \underline{102.3}  & \underline{18.3}   \\
        \ding{174}     &   \checkmark         & \checkmark  & \textbf{105.3}  & \textbf{18.8} \\ \bottomrule
    \end{tabular}
    \caption{Ablation on the output of BSIM.}
    \vspace{-2.0em}
    \label{table:example6}
\end{table}

Model~\ding{174} extends Model~\ding{172} by incorporating the PIN, significantly boosting performance (CIDEr: 66.5 → 95.1). Based on this, model~\ding{175} further improves performance by leveraging key video content. Video content provides details on player actions and positions. Integrating this with player identities helps the model accurately associate players with their actions, leading to more precise descriptions. As a result, Model~\ding{176} significantly outperforms Model~\ding{174}. Model~\ding{177} integrates all components, achieving the best performance. These results confirm the effectiveness of each module, highlighting the crucial role of visual-based player identification in identity-aware sports video captioning. We further explore the impact of the BSIM module outputting only a single enhanced feature (either video or player) on model performance. In \cref{table:example6}, BSIM performs worse when outputting only one enhanced feature compared to using both. The enhanced player features contribute the most to overall performance.

\section{Conclusion, Limitation and Future Work}
\vspace{-5pt}
We propose a player-centric multimodal prompt generation network for identity-aware basketball video captioning. An identity-related information extraction module is designed to help the LLM recognize player identities and generate descriptions with player names. To support this work, we construct a large identity-aware basketball video captioning dataset containing 9,726 videos, 321 players with bounding boxes, and 9 major event types. Extensive experiments validate the effectiveness of our method.

As an initial exploration of player recognition for identity-aware sports video captioning, our work still has some limitations. Firstly, the model heavily relies on the performance of the player identification network. Future work should fully utilize the player-related knowledge, such as player numbers and jersey colors, to improve identity recognition accuracy. Secondly, the model struggles to infer shot distances, and the distance evaluation module should be added to address this issue. Thirdly, the dataset needs to include more event types to make it more comprehensive. Despite these limitations, we believe that recognizing player identities from a visual perspective is a key breakthrough for identity-aware sports video captioning and offers valuable insights. Furthermore, the proposed dataset provides valuable resources for future research.

\section*{Acknowledgments}
This project was supported in part by the Beijing Natural Science Foundation under grant L233008; in part by the Natural Science Foundation of China under Grant 62236010.

%% file: appendix.tex
\clearpage
\setcounter{page}{1}
\maketitlesupplementary
\appendix
\section{Identity-aware Basketball Video Captioning Dataset}
\label{sec:Identity-aware Basketball Video Captioning Dataset}

With the improvement of people's living standards, the vigorous development of sports undertakings, and the growing enthusiasm of the public for sports, the demand for sports video understanding is also expanding. In sports video captioning, there are typically two types of descriptions: 1) commentary, which targets TV audiences and tends to be long and detailed; 2) live text broadcasting, which targets online users and is more concise. Our work focuses on the latter, where the captions are brief but include key events with corresponding players. Moreover, with the fast pace of life and busy work schedules, people may not have the time to watch sports videos. Compared to videos, which require people to spend a certain amount of time to understand the game situation, text live broadcasts can provide people with concise and clear text, allowing them to quickly grasp the information of the videos. Therefore, for basketball live text broadcast, we aim to construct a dataset whose text provides concise event descriptions, including actions and participants, helping audiences quickly grasp key content and ensuring efficient updates and summaries for sports fans. 

The proposed NBA-Identity dataset is designed for basketball live text broadcasting. Each video clip of an event is annotated with a corresponding description. Unlike traditional video captioning datasets that summarize video content with broad overviews, this dataset focuses on providing descriptions with specific player identities and actions to describe visual content. This section introduces the following aspects of the NBA-Identity dataset: 1) dataset collection process, 2) statistical analysis, and 3) dataset versatility.

\subsection{Dataset Collection}
\begin{figure}[t!]
\centering
\includegraphics[width=8.4cm,height=5.0cm]{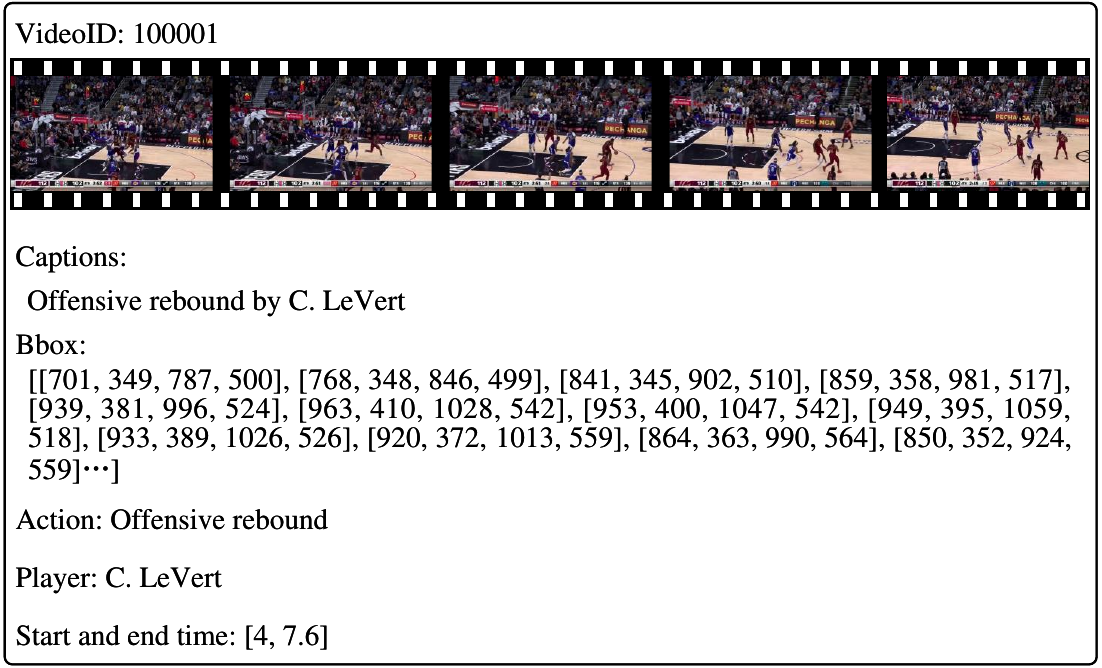}
	\caption{Data samples from the proposed dataset. Each video clip is annotated by video\_id, caption, bounding box, action type and player names.}
	\label{fig:supp1}
    \vspace{-1.5em}
\end{figure}

We collect text data from 40 games through the professional basketball data platform and obtain corresponding video data from a basketball live streaming platform. Using the methodology of VC-NBA-2022~\cite{xi2024knowledge}, the text and video data are aligned through Tesseract-OCR~\cite{smith2007overview}. The event types we used are sourced from professional basketball live broadcast data. A few low-frequency event types, such as ``jump ball'' (about 1/350 of all events per game) and ``violation'' (about 1/350 of all events per game), are excluded. From these 40 games, we extract a total of 9,726 video clips, covering 321 player identities and 9 major types of events: ``block'', ``foul'', ``defensive rebound'', ``offensive rebound'', ``turnover'', ``two-point (2-pt) shot'', ``three-point (3-pt) shot'', ``layup'', and ``assist''. The nine major event types mentioned in the paper are broad categories, but for example, ``foul'' includes ``personal foul'', ``personal take foul'', ``shooting foul'', ``offensive foul'', and ``loose ball foul''. Similarly, ``turnover'' includes ``lost ball; steal by [player]'', ``bad pass; steal by [player]'', ``lost ball'', ``bad pass'', ``offensive foul'', ``traveling'', ``step out of bounds'', ``discontinued dribble'', and ``out of bounds lost ball''. ``Shot'' events are further divided into ``make shot'' and ``miss shot''. The dataset also contains events with limited samples, such as ``free throw'' and ``drawn''. Our dataset actually covers a total of 26 fine-grained event types in basketball. Notably, the VC-NBA-2022 dataset discards shooting distance (\eg, ``from 20 ft'') and combines missed shots with rebounds into a single event (\eg, ``L. Shamet misses the 3pt jump shot and E. Gordon gets the defensive rebound''). In contrast, we retain the shooting distance to enhance the dataset's complexity and authenticity. The missed shots and rebounds are not merged because it could lead to increased video duration, and the rapid transitions of scenes would make the identification of these events easier. Furthermore, we annotate the coordinates of the key players for each video clip. Key players refer to the players included in the description of the video clip. The dataset sample is shown in~\cref{fig:supp1}.  NBA-Identity is split by game instances, with 35 games (8,667 clips) for training and 5 games (1,059 clips) for testing.

\begin{figure*}[t!]
    \centering
    \begin{tabular}{@{\extracolsep{\fill}}c@{}c@{\extracolsep{\fill}}c@{}c@{\extracolsep{\fill}}c@{}c@{\extracolsep{\fill}}}
            \includegraphics[width=8.6cm,height=5.2cm]{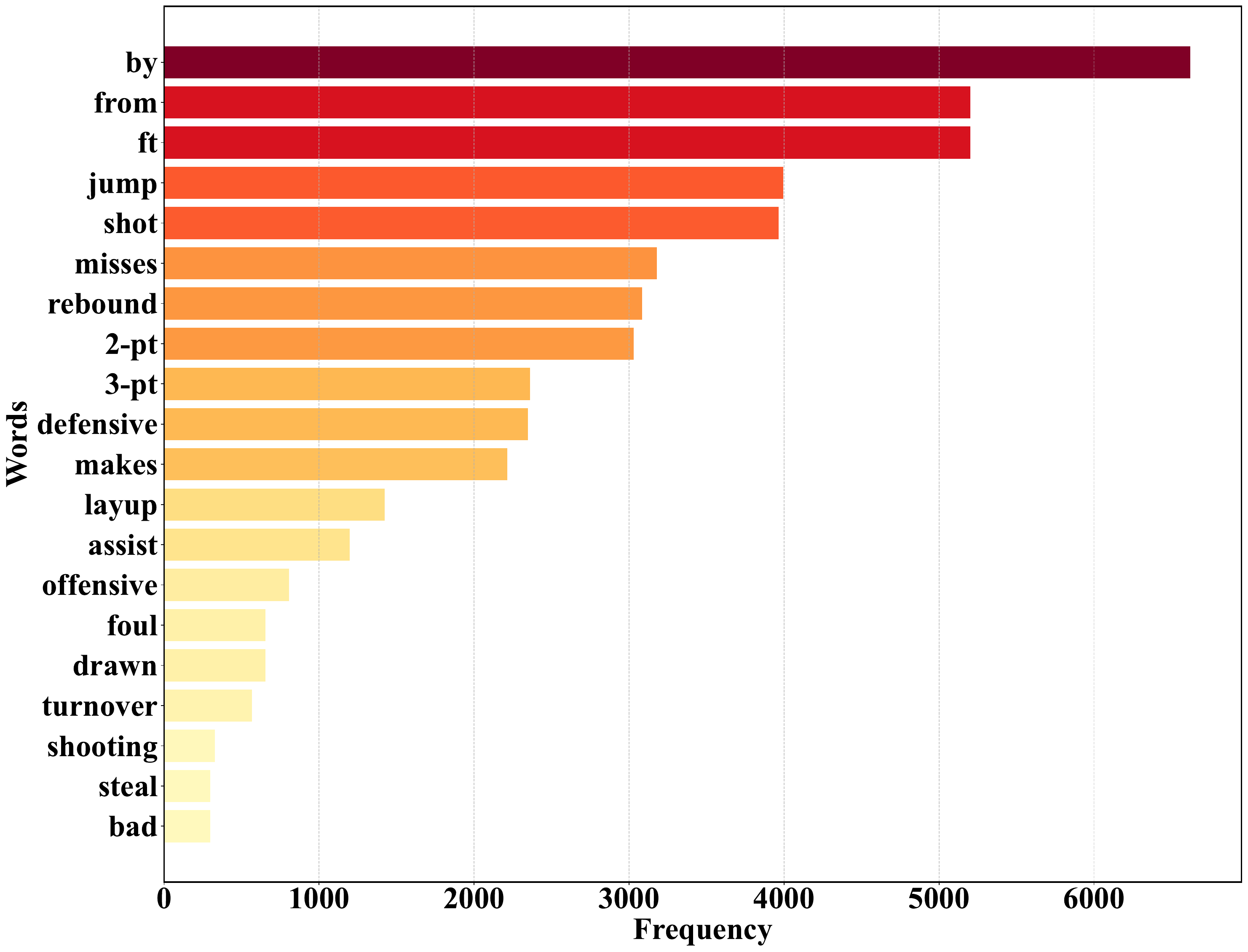} &
            \includegraphics[width=8.6cm,height=5.2cm]{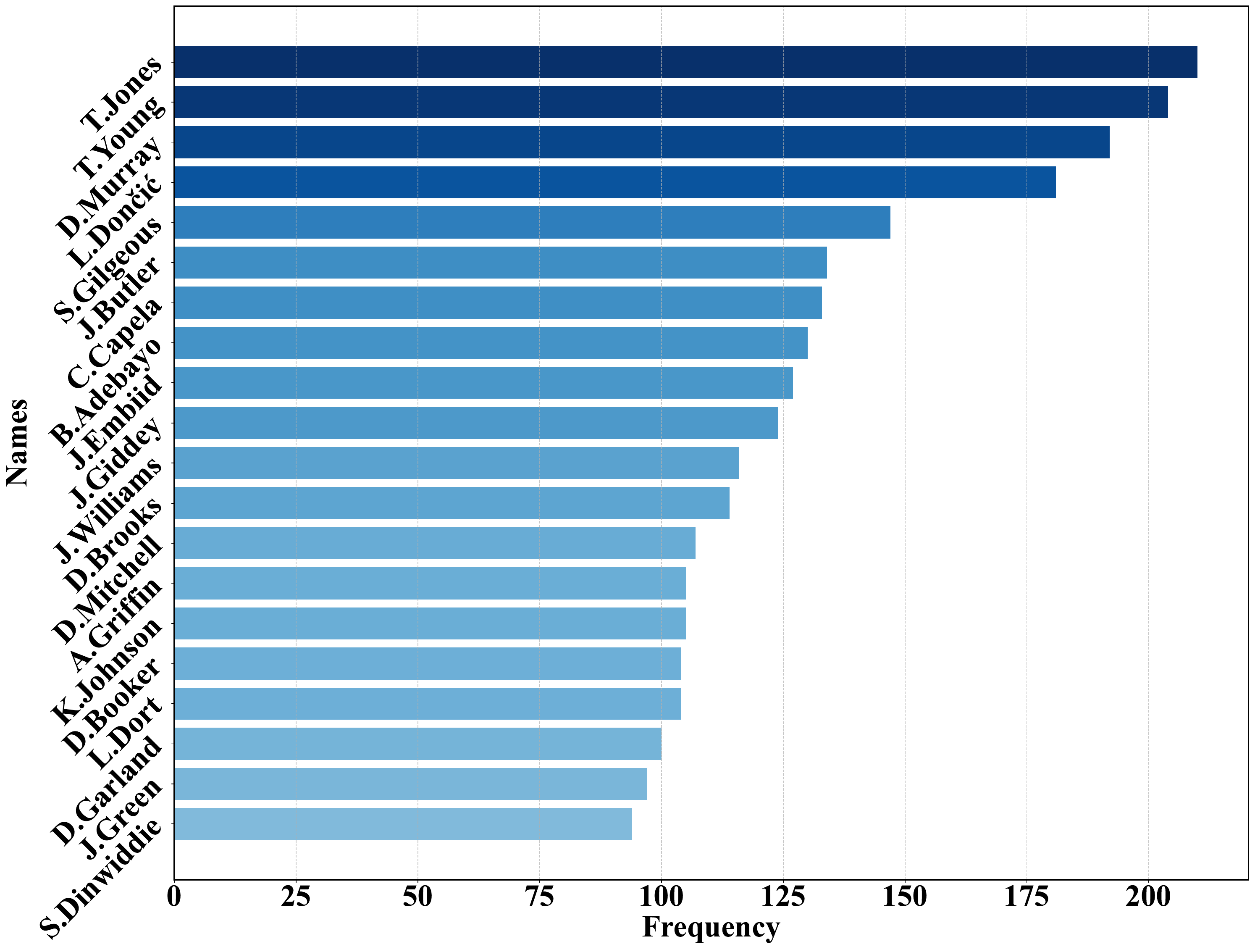}
            \\
            (a) Distribution of frequent words & (b) Distribution of frequent names \\
    \end{tabular}

    \caption{Illustrations of NBA-Identity dataset statistics.}
    \label{fig:supp2}
\end{figure*}

\begin{figure}[t!]
\centering
\includegraphics[width=8.2cm,height=6.0cm]{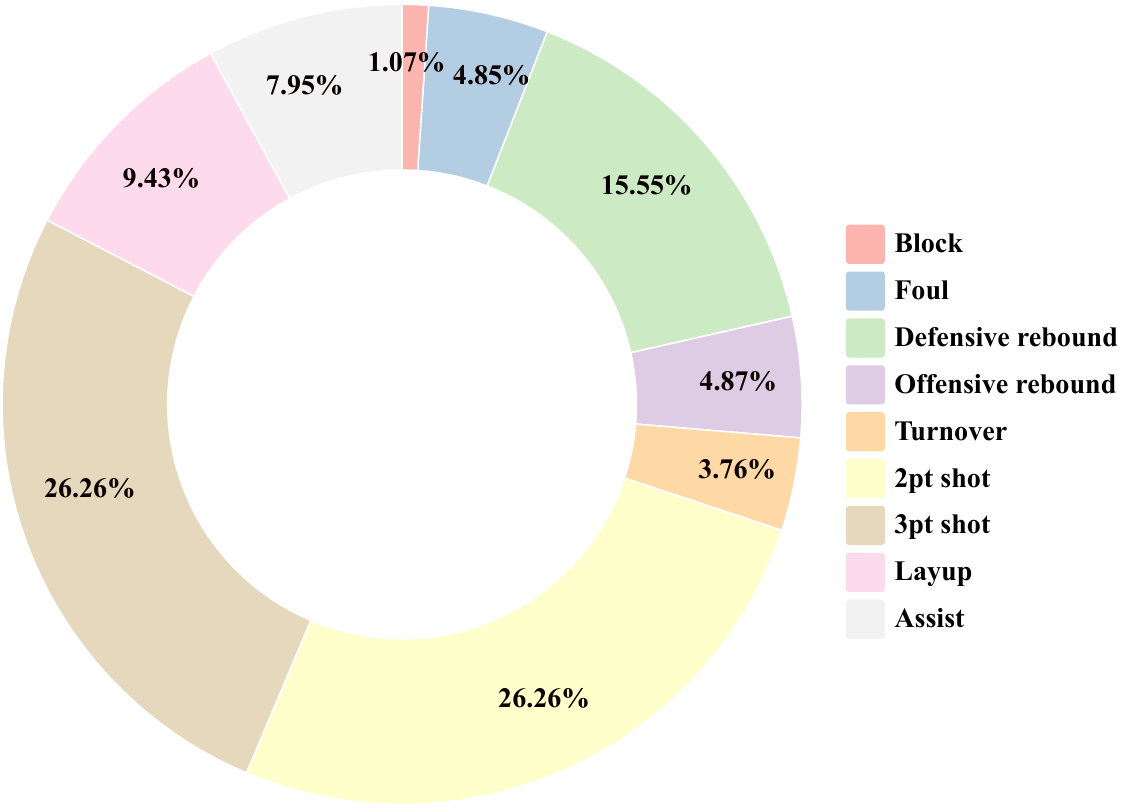}
	\caption{The distribution of basketball action categories contained in NBA-Identity dataset.}
	\label{fig:supp3}
\end{figure}
\begin{figure}[t!]
\centering
\includegraphics[width=8.2cm,height=6.0cm]{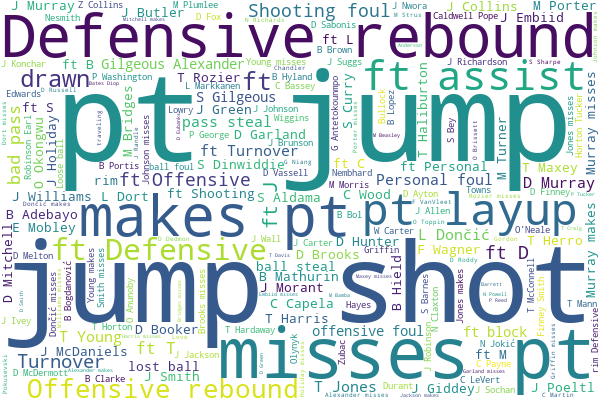}
	\caption{Word cloud of NBA-Identity dataset. The bigger the font, the more percentage it occupies.}
	\label{fig:supp4}
\end{figure}
\begin{figure}[t]
    \centering
    \includegraphics[width=8.4cm,height=2.8cm]{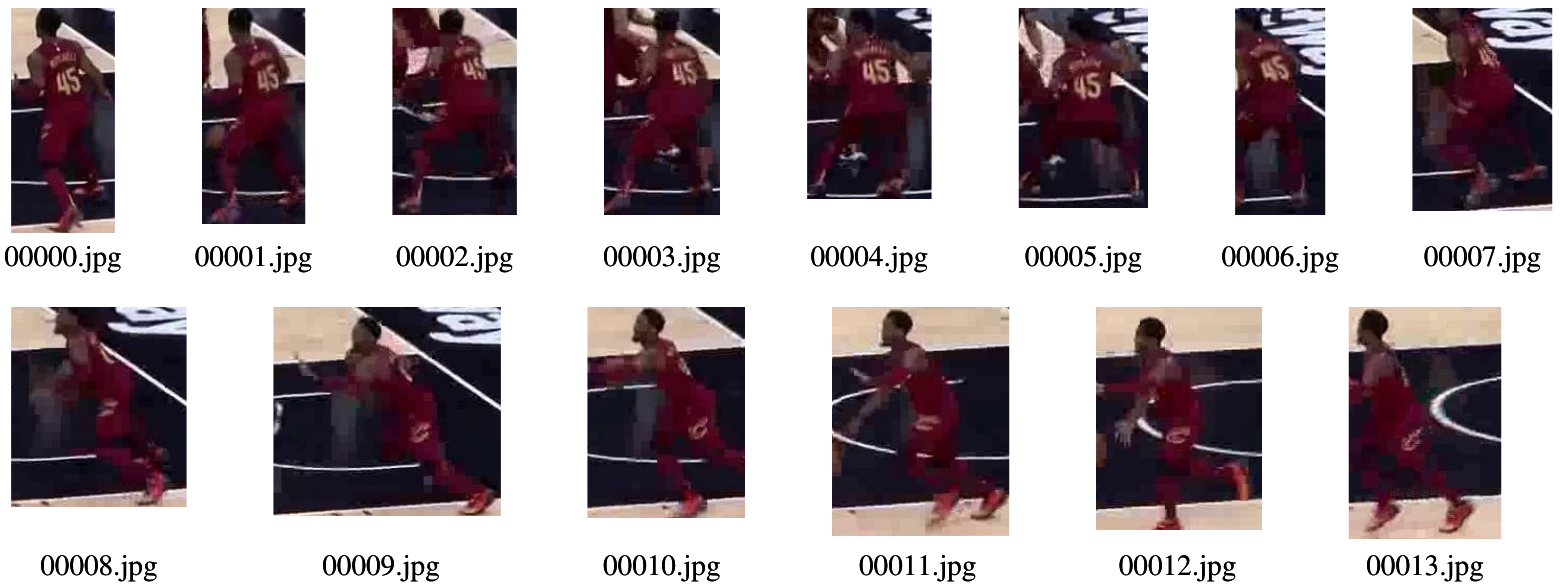}
    \caption{Example of player sequence.}
    \label{fig:supp5}
\end{figure}

\begin{figure*}[t!]
    \centering
    \includegraphics[width=17.4cm,height=14.4cm]{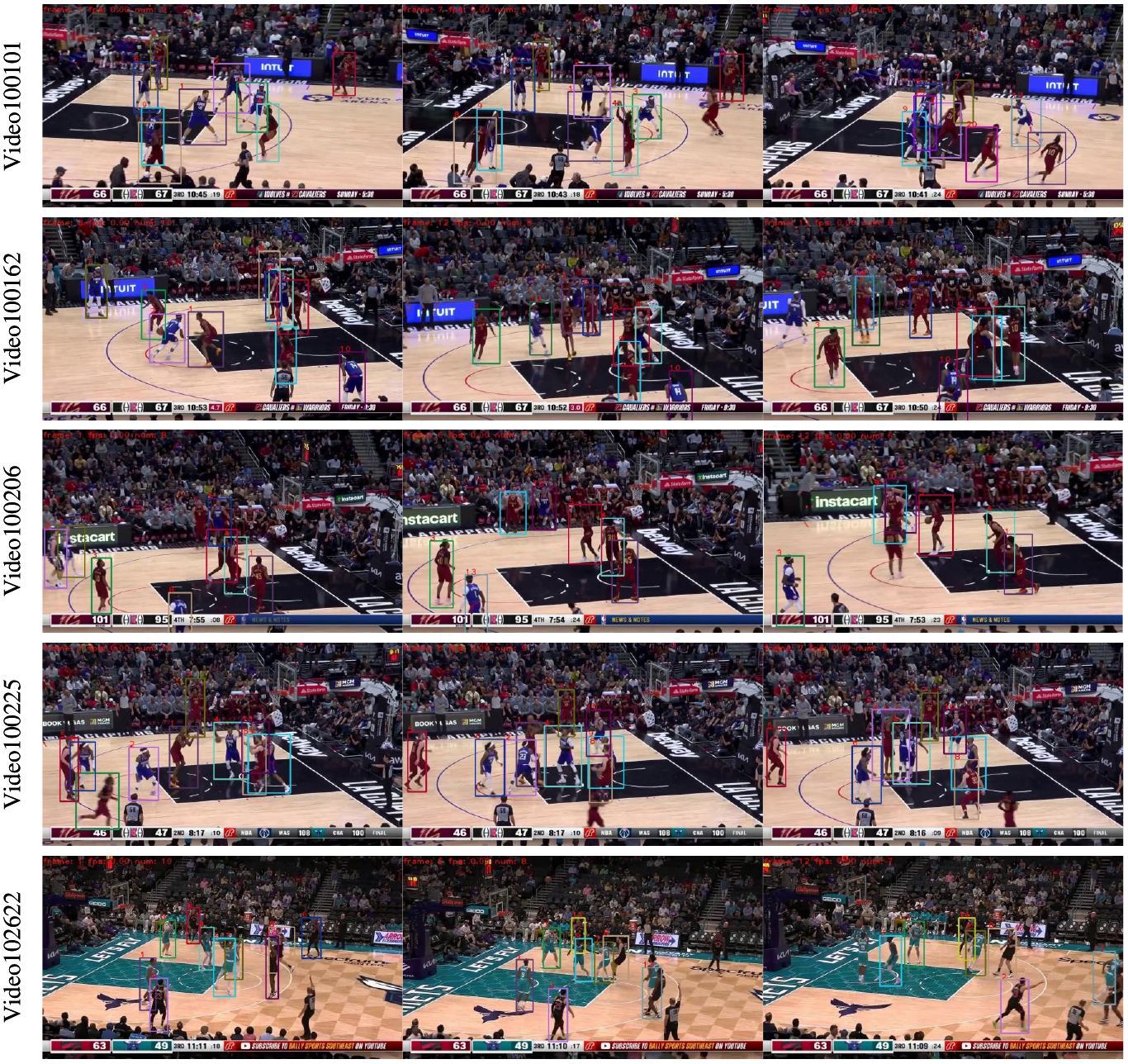}
    \caption{Example tracking results of SportsMOT on the test set of NBA-Identity. Each row shows the results of sampled frames in chronological order of a video clip. Bounding boxes and players are marked in the images. Bounding boxes with different colors represent different players.}
    \label{fig:supp6}
\end{figure*}

\subsection{Dataset Statistics}
Our dataset is the largest identity-aware video captioning dataset in sports domain, both in terms of the number of videos and descriptions, with a particular emphasis on annotating player identities. This provides researchers with a richer data resource to support in-depth analysis and understanding of basketball events. The distribution of the top 20 most frequent words in the dataset is shown in \cref{fig:supp2} (a), illustrating common descriptive vocabulary. \cref{fig:supp2} (b) presents the frequency of player identities, with the highest occurrence being that of T.Jones. These statistics offer insights into the trends in descriptions and player involvement, providing valuable references for subsequent modeling efforts. Additionally, \cref{fig:supp3} illustrates the distribution of different event types in the dataset. Shooting and rebounding events occur most frequently, while blocking events are the least common. This distribution aligns with real-game scenarios, reflecting the typical frequency of various events in basketball games. We also provide word-cloud-based statistics in \cref{fig:supp4} to reveal the relative amount of different words. It shows that the top-5 subjects in NBA-Identity are ``jump'', ``shot'', ``pt'', ``misses'' and ``defensive'', followed by ``rebound'', ``makes'', ``layup'' and ``defensive''.

\subsection{Dataset Versatility}
This dataset demonstrates significant versatility and development potential, offering a valuable resource for future research. In addition to providing detailed descriptions for each video clip, the dataset includes annotations for action types, player coordinates, and the temporal boundaries of events. These comprehensive annotations extend the dataset's application range, supporting not only video captioning tasks but also enabling research in group activity recognition~\cite{wu2024learning}, player identity recognition~\cite{vats2023player} and temporal action detection~\cite{hu2024overview}. Furthermore, the multidimensional annotation details offer researchers opportunities to explore correlations between video data and event characteristics, laying a strong foundation for in-depth research in the field of sports analysis.
\vspace{-0.5em}

\section{Evaluation Metrics}
We employ typical captioning metrics (\eg, BLEU (B)~\cite{papineni2002bleu}, Rouge-L (R)~\cite{lin2004rouge}, METEOR (M)~\cite{banerjee2005meteor} and CIDEr (C)~\cite{vedantam2015cider}) to evaluate the performance of LLM-IAVC. 

(1) BLEU is one of the most widely used metrics in machine translation and captioning. It evaluates the quality of generated captions by calculating the precision of n-gram matches between the generated and reference captions, typically using BLEU-4 (4-gram) as the standard.
\begin{equation}
\mathrm{BLEU} = \mathrm{BP} \cdot \mathrm{exp}\left ( \sum_{N}^{n=1}  \mathrm{log} p_{n}   \right ),
\label{eq:equation15}
\end{equation}
where $p_{n}$ is the n-gram precision, calculated as the number of matched n-grams divided by the total number of n-grams in the generated caption. $\mathrm{BP}$ denotes the Brevity Penalty, which penalizes overly short generated captions:

\begin{equation}
\mathrm{BP} = \begin{cases}
1 ~~~~~~~~~~~if~ c>r  
\\
e^{1-r/c}~~~ if~c\le r
\end{cases},
\label{eq:equation16}
\end{equation}
where $c$ denotes the length of the generated caption and $r$ denotes the length of the shortest reference caption. $w_{n}$ is the weight for n-grams, typically set to $w_{n} =\frac{1}{N} ~\left ( e.g. N=4~for~\mathrm{BLEU-4}  \right ) $.

(2) Rouge-L evaluates the quality of generated captions by calculating the longest common subsequence (LCS) between the generated and reference captions, with a focus on recall ($\mathrm{R}$). 
\begin{equation}
\mathrm{R_{LCS}} =\frac{\mathrm{LCS\left ( \textbf{X},\textbf{Y} \right )} }{\mathrm{len(\textbf{Y})}},
\label{eq:equation17}
\end{equation}
\begin{equation}
\mathrm{P_{LCS}} =\frac{\mathrm{LCS\left ( \textbf{X},\textbf{Y} \right )} }{\mathrm{len(\textbf{X})}},
\label{eq:equation18}
\end{equation}
\begin{equation}
\mathrm{Rouge-L}=\frac{\left ( 1+\beta ^{2} \right )\mathrm{R_{LCS}P_{LCS}}  }{\mathrm{R_{LCS}}+\beta ^{2}\mathrm{P_{LCS}}},  
\label{eq:equation19}
\end{equation}
where $\mathrm{\textbf{X}}$ and  $\mathrm{\textbf{Y}}$ denote the generated caption and the reference caption, respectively. $\mathrm{LCS\left ( \textbf{X},\textbf{Y} \right )}$ demotes the length of the longest common subsequence between $\mathrm{\textbf{X}}$ and  $\mathrm{\textbf{Y}}$. And $\beta$ is the weight parameter, typically set to 1 (balancing recall and precision). $\mathrm{P}$ is the precision score.

(3) METEOR combines precision ($\mathrm{P}$), recall ($\mathrm{R}$), and word order in its evaluation, closer to human evaluation. It incorporates synonym matching and stemming to improve robustness.
\begin{equation}
\mathrm{METEOR}=\left ( 1-\gamma \cdot \mathrm{Penalty} \right ) \cdot \frac{\mathrm{P}\cdot \mathrm{R}}{\alpha \mathrm{P}+\left ( 1-\alpha  \right )\mathrm{R} } ,
\label{eq:equation20}
\end{equation}
where $\mathrm{P}$ denotes the precision score, calculated as the number of matched words divided by the total number of words in the generated caption. $\mathrm{R}$ denotes the recall score, calculated as the number of matched words divided by the total number of words in the reference caption. $\alpha$ is the weight parameter for balancing precision and recall, typically set to 0.9. $\gamma$ is the the penalty weight, typically set to 0.5. And $\mathrm{Penalty}$ denotes the word order penalty, calculated based on the difference in word order between the generated and reference captions.

(4) CIDEr is specially designed for captioning task evaluation. It evaluates the semantic consistency of generated captions by computing the similarity between the generated and reference captions using TF-IDF weighting.
\begin{equation}
\mathrm{CIDEr\left ( \textbf{C}, \textbf{S} \right )} =\frac{1}{m} \sum_{m}^{j=1}\frac{\mathrm{g\left ( \textbf{C} \right )}\cdot \mathrm{ g\left ( \textbf{s}_{j}  \right )}  }{\left \|\mathrm{g\left ( \textbf{C} \right )}  \right \|\cdot\left \| \mathrm{g\left ( \textbf{s}_{j}  \right )} \right \| }   ,
\label{eq:equation21}
\end{equation}
where $\mathrm{\textbf{C}}$ denotes the generated caption. $\mathrm{\textbf{S}}=\left \{ \mathrm{\textbf{s}}_{1} ,\mathrm{\textbf{s}}_{2},...,\mathrm{\textbf{s}}_{m}   \right \} $ denotes the set of reference captions. $\mathrm{g\left ( \cdot \right )}$ denotes the TF-IDF vector representation of a caption. The cosine similarity between the generated caption and each reference caption is calculated and averaged.

\section{SportsMOT: Multi-object Tracker}
At the training stage, player sequences are cropped from the video based on the bounding boxes provided in the dataset, as shown in~\cref{fig:supp5}. Subsequently, the player recognition module extracts visual features from these sequences for further processing. However, providing player bounding boxes directly during the inference stage is unreasonable and impractical. Therefore, at the inference stage, the multi-objective tracker SportsMOT~\cite{cui2023sportsmot} is utilized to extract player sequences of multiple players. It focuses on multi-player tracking and excludes audience members and referees. Our proposed player identification network extracts their visual features and obtain corresponding player names from player sequences. The multi-object tracking results are shown in~\cref{fig:supp6}. 

\begin{table}[t!]
\footnotesize
    \centering

\begin{tabular}{@{}l|c|cc@{}}
\toprule
Model                                 & Down\_dim & CIDEr & BLEU-4 \\ \midrule
\multirow{4}{*}{LLM-IAVC (Llama3.2-3B)} & 128      & 99.6  & 15.8   \\
                                      & 256      & 100.1 & 17.2   \\
                                      & 512      & 105.3 & 18.8   \\
                                      & 768      & 105.3 & 18.0   \\ \bottomrule
\end{tabular}%
\caption{Impact of dimension settings in down-projection matrix.}
\label{tab:example6}
\end{table}
\begin{figure}[t!]
    \centering
    \includegraphics[width=8.0cm,height=4.8cm]{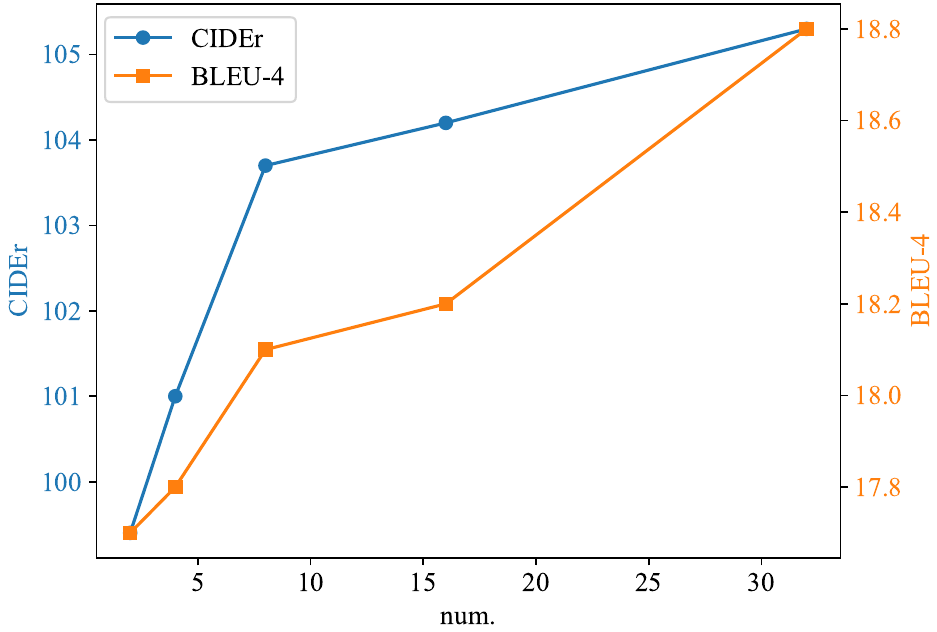}
    \caption{Ablation on the number of learnable vectors in VCLM.}
    
    \label{fig:supp7}
\end{figure}

\section{Additional Ablation Studies}
\noindent \textbf{Effects of Different Dimensions of BSIM}. BSIM initially compresses the input features into a lower dimension, subsequently performs interaction operations, and ultimately reconstructs the dimensions back to their original size. This approach improves computational efficiency by reducing redundant information. It also ensures that the interaction emphasizes the most important features, which in turn enhances the model's performance and generalization ability. We also conduct experiments comparing the BSIM module’s performance with different intermediate lower dimensions. As shown in \cref{tab:example6}, when the dimension is set to 512, the model achieves optimal performance, with CIDEr at 105.3 and BLEU-4 at 18.8. This design allows BSIM to retain essential information while reducing unnecessary computational overhead, ultimately improving the overall efficiency and effectiveness of the model.

\noindent \textbf{Ablation on the number of learnable vectors in VCLM}. We exploit how the number of learnable query vectors affects LLM-IAVC performance. As shown in \cref{fig:supp7}, the model performs best with 32 vectors. Performance declines when the number is below 18, as fewer tokens capture less key video content. Conversely, using more than 32 vectors introduces redundant information. Ultimately, the number of query vectors for VCLM is set to 32.

\section{Discussion}
\noindent \textbf{Dataset}. Compared to existing identity-aware sports video captioning datasets, our dataset contains the largest number of videos. Each visual scene varies with changes in lighting, player positions, and the appearance of actions. Given the high annotation cost, like other sports video captioning datasets, each video has only one description. These captions crawled from live text commentary websites primarily include different player names, actions, and distances. In addition, captions on our dataset aim to help audiences quickly understand match information (specifically which player performs what action). As a result, the caption does not need to prioritize diversity, creativity, or rich expressions. Each action is described using official and professional terminology.

\noindent \textbf{The generality of pipeline}. Our method is a highly generalizable framework that can be seamlessly applied to a wide range of sports, including but not limited to volleyball, table tennis, baseball, and soccer, without requiring any modifications to its core modules. However, it requires manually annotating player bounding boxes to pre-train the corresponding player identification network for each sport. This step is crucial for enabling the model to accurately identify and track players within the unique visual contexts and dynamics of each sport.